\documentclass[11pt]{article}
\usepackage{acl2016}
\usepackage{graphicx}
\usepackage{color}
\usepackage{flushend}
\usepackage{mathtools}
\usepackage{epstopdf}

\usepackage{varwidth} 
\usepackage{url,graphicx,times}
\usepackage{tabularx,amsmath}
\usepackage{amssymb}
\usepackage{multirow}
\usepackage{booktabs}
\usepackage{enumerate}
\usepackage{subfigure}
\usepackage{enumitem}
\usepackage{setspace}
\usepackage{verbatim}

\date{}

\begin{document}

\aclfinalcopy

\title{Recurrent Dropout without Memory Loss}

\author{
	    Stanislau Semeniuta\\
	    Universit{\"a}t zu L{\"u}beck\\
	    Institut f{\"u}r Neuro- und Bioinformatik \\
	    {\tt stas@inb.uni-luebeck.de}\\
	  \And
		Aliaksei Severyn\\
	    Google Inc.\\
	    {\tt severyn@google.com}
	  \And
		Erhardt Barth \\
	    Universit{\"a}t zu L{\"u}beck\\
	    Institut f{\"u}r Neuro- und Bioinformatik \\
	    {\tt barth@inb.uni-luebeck.de}\\
}

\author{Stanislau Semeniuta$^{1}$ \; Aliaksei Severyn$^{2}$ \; Erhardt Barth$^{1}$\\
$^{1}$Universit{\"a}t zu L{\"u}beck, Institut f{\"u}r Neuro- und Bioinformatik \\
 \normalsize{\tt \{stas,barth\}@inb.uni-luebeck.de}\\
$^{2}$Google Research \\
 \normalsize{\tt severyn@google.com}\\
}

\maketitle

\begin{abstract}
This paper presents a novel approach to recurrent neural network (RNN) regularization. Differently from the widely adopted dropout method, which is applied to \textit{forward} connections of feed-forward architectures or RNNs, we propose to drop neurons directly in \textit{recurrent} connections in a way that does not cause loss of long-term memory. Our approach is as easy to implement and apply as the regular feed-forward dropout and we demonstrate its effectiveness for Long Short-Term Memory network, the most popular type of RNN cells. Our experiments on NLP benchmarks show consistent improvements even when combined with conventional feed-forward dropout. 
\end{abstract}


\section{Introduction}
\label{sec:intro}
Recurrent Neural Networks, LSTMs in particular, have recently become a popular tool among NLP researchers for their superior ability to model and learn from sequential data. These models have shown state-of-the-art results on various public benchmarks ranging from sentence classification~\cite{lstm-acl15-tweet,lstm-emnlp14-opinion,lstm-acl15-doc} and various tagging problems~\cite{lstm-acl15-parsing} to language modelling~\cite{rec-lm,rec-lm2}, text generation~\cite{lstm-emnlp14-generation} and sequence-to-sequence prediction tasks~\cite{seq2seq}.



Having shown excellent ability to capture and learn complex linguistic phenomena, RNN architectures are prone to overfitting. Among the most widely used techniques to avoid overfitting in neural networks is the dropout regularization~\cite{hinton_dropout}. Since its introduction it has become, together with the L2 weight decay, the standard method for neural network regularization. While showing significant improvements when used in feed-forward architectures, e.g., Convolutional Neural Networks~\cite{imagenet_krizhevsky}, the application of dropout in RNNs has been somewhat limited. Indeed, so far dropout in RNNs has been applied in the same fashion as in feed-forward architectures: it is typically injected in input-to-hidden and hidden-to-output connections, i.e., along the input axis, but not between the recurrent connections (time axis). Given that RNNs are mainly used to model sequential data with the goal of capturing short- and long-term interactions, it seems natural to also regularize the recurrent weights. This observation has led us and other researchers \cite{moon2015rnndrop,gal2015dropout} to the idea of applying dropout to the recurrent connections in RNNs. 

In this paper we propose a novel \textit{recurrent dropout} technique and demonstrate how our method is superiour to other recurrent dropout methods recently proposed in~\cite{moon2015rnndrop,gal2015dropout}. Additionally, we answer the following questions which helps to understand how to best apply recurrent dropout:
(i) how to apply the dropout in recurrent connections of the LSTM architecture in a way that prevents possible corruption of the long-term memory;
(ii) what is the relationship between our \textit{recurrent dropout} and the widely adopted dropout in input-to-hidden and hidden-to-output connections;
(iii) how the dropout mask in RNNs should be sampled: once per step or once per sequence. 
The latter question of sampling the mask appears to be crucial in some cases to make the recurrent dropout work and, to the best of our knowledge, has received very little attention in the literature. Our work is the first one to provide empirical evaluation of the differences between these two sampling approaches.

Regarding empirical evaluation, we first highlight the problem of information loss in memory cells of LSTMs when applying \textit{recurrent dropout}. We demonstrate that previous approaches of dropping \textit{hidden state} vectors cause loss of memory while our proposed method to use dropout mask in \textit{hidden state update} vectors does not suffer from this problem. We experiment on three widely adopted NLP tasks: word- and character-level Language Modeling and Named Entity Recognition. The results demonstrate that our \textit{recurrent dropout} helps to achieve better regularization and yields improvements across all the tasks, even when combined with the conventional feed-forward dropout. Furthermore, we compare our dropout scheme with the recently proposed alternative recurrent dropout methods and show that our technique is superior in almost all cases.




\section{Related Work}
\label{sec:related-work}
Neural Network models often suffer from overfitting, especially when the number of network parameters is large and the amount of training data is small. This has led to a lot of research directed towards improving their generalization ability. Below we primarily discuss some of the methods aimed at improving regularization of RNNs.

\newcite{DBLP:journals/corr/PhamKL13} and \newcite{DBLP:journals/corr/ZarembaSV14}  have shown that LSTMs can be effectively regularized by using dropout in forward connections. While this already allows for effective regularization of recurrent networks, it is intuitive that introducing dropout also in the hidden state may force it to create more robust representations. Indeed, \newcite{moon2015rnndrop} have extended the idea of dropping neurons in forward direction and proposed to drop cell states as well showing good results on a Speech Recognition task. \newcite{Bluche2015b} carry out a study to find where dropout is most effective, e.g. input-to-hidden or hidden-to-output connections. The authors conclude that it is more beneficial to use it once in the correct spot, rather than to put it everywhere. \newcite{DBLP:journals/corr/BengioVJS15} have proposed an algorithm called scheduled sampling to improve performance of recurrent networks on sequence-to-sequence labeling tasks. A disadvantage of this work is that the scheduled sampling is specifically tailored to this kind of tasks, what makes it impossible to use in, for example, sequence-to-label tasks. \newcite{gal2015dropout} uses insights from variational Bayesian inference to propose a variant of LSTM with dropout that achieves consistent improvements over a baseline architecture without dropout. 

The main contribution of this paper is a new \textit{recurrent dropout} technique, which is most useful in gated recurrent architectures such as LSTMs and GRUs. 
We demonstrate that applying dropout to arbitrary vectors in LSTM cells may lead to loss of memory thus hindering the ability of the network to encode long-term information. In other words, our technique allows for adding a strong regularizer on the model weights responsible for learning short and long-term dependencies without affecting the ability to capture long-term relationships, which are especially important to model when dealing with natural language. Finally, we compare our method with alternative \textit{recurrent dropout} methods recently introduced in~\cite{moon2015rnndrop,gal2015dropout} and demonstrate that our method allows to achieve better results.

\section{Recurrent Dropout}
\label{sec:model}
In this section we first show how the idea of feed-forward dropout~\cite{hinton_dropout} can be applied to recurrent connections in vanilla RNNs. We then introduce our \textit{recurrent dropout} method specifically tailored for gated architectures such as LSTMs and GRUs. We draw parallels and contrast our approach with alternative recurrent dropout techniques recently proposed in~\cite{moon2015rnndrop,gal2015dropout} showing that our method is favorable when considering potential memory loss issues in long short-term architectures.

%

\subsection{Dropout in vanilla RNNs}

Vanilla RNNs process the input sequences as follows:

\begin{equation}\label{eq:rnn}
\mathbf{h_t} = f(\mathbf{W_{h}} [\mathbf{x_t},\mathbf{h_{t-1}}] + \mathbf{b_h}),
\end{equation}	

\noindent where $\mathbf{x_t}$ is the input at time step $t$; $\mathbf{h_t}$ and $\mathbf{h_{t-1}}$ are hidden vectors that encode the current and previous states of the network; $\mathbf{W_{h}}$ is parameter matrix that models input-to-hidden and hidden-to-hidden (recurrent) connections; $\mathbf{b}$ is a vector of bias terms, and \(f\) is the activation function.

As RNNs model sequential data by a fully-connected layer, dropout can be applied by simply dropping the previous hidden state of a network. Specifically, we modify Equation~\ref{eq:rnn} in the following way:
\begin{equation}
\mathbf{h_t} = f(\mathbf{W_{h}} [\mathbf{x_t},d(\mathbf{h_{t-1}})] + \mathbf{b_h}),
\end{equation}
\noindent where \(d\) is the dropout function defined as follows:  
\begin{equation}\label{eq:dropout}
d(\mathbf{x}) = \begin{dcases}
mask * \mathbf{x},& \text{if train phase}\\
(1-p)\mathbf{x}       & \text{otherwise,}
\end{dcases}
\end{equation}
\noindent where \(p\) is the dropout rate and \(mask\) is a vector, sampled from the Bernoulli distribution with success probability \(1-p\). 
\subsection{Dropout in LSTM networks}
\begin{figure*}
	\centering
	\subfigure[Moon et al., 2015]{\centering\includegraphics[height=0.15\textwidth]{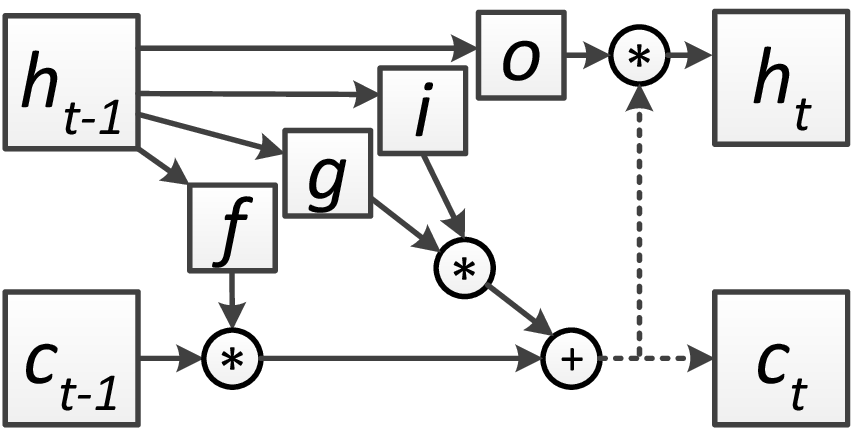}}
	\quad
	\subfigure[Gal, 2015]{\centering\includegraphics[height=0.15\textwidth]{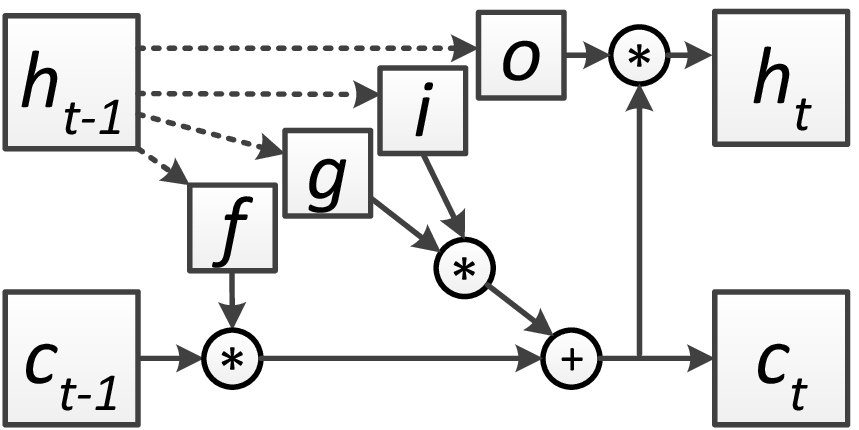}}
	\quad
	\subfigure[Ours]{\centering\includegraphics[height=0.15\textwidth]{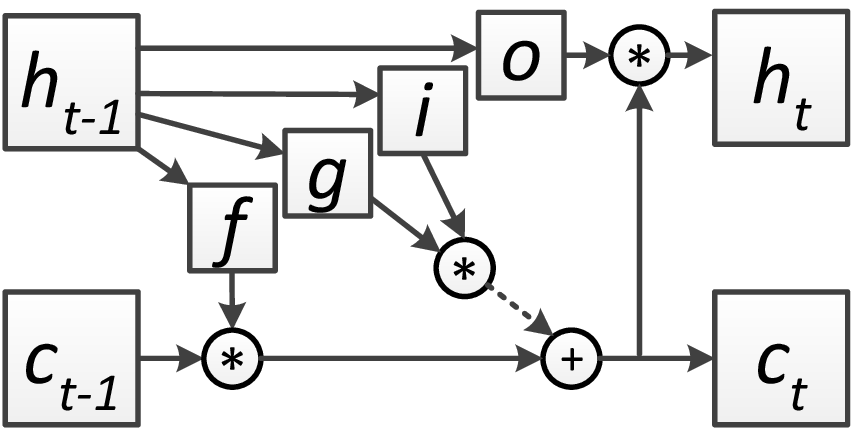}}
	\caption{Illustration of the three types of dropout in recurrent connections of LSTM networks. Dashed arrows refer to dropped connections. Input connections are omitted for clarity.}
	\label{fig:networks}
\end{figure*}



Long Short-Term Memory networks~\cite{Hochreiter:1997:LSM:1246443.1246450} have introduced the concept of gated inputs in RNNs, which effectively allow the network to preserve its memory over a larger number of time steps during both forward and backward passes, thus alleviating the problem of vanishing gradients~\cite{Bengio-trnn94}. Formally, it is expressed with the following equations:

\begin{equation}\label{eq:lstm_candidates}
\begin{pmatrix}
\mathbf{i}_{t} \\
\mathbf{f}_{t} \\
\mathbf{o}_{t} \\
\mathbf{g}_{t} \\ 
\end{pmatrix}
=
\begin{pmatrix}
\sigma(\mathbf{W}_{i} \begin{bmatrix}\mathbf{x}_t , \mathbf{h}_{t-1} \end{bmatrix} + \mathbf{b}_i)\\
\sigma(\mathbf{W}_{f} \begin{bmatrix}\mathbf{x}_t , \mathbf{h}_{t-1} \end{bmatrix} + \mathbf{b}_f) \\
\sigma(\mathbf{W}_{o} \begin{bmatrix}\mathbf{x}_t , \mathbf{h}_{t-1} \end{bmatrix} + \mathbf{b}_o) \\
f(\mathbf{W}_{g} \begin{bmatrix}\mathbf{x}_t , \mathbf{h}_{t-1} \end{bmatrix} + \mathbf{b}_g)
\end{pmatrix}
\end{equation}
\begin{equation}\label{eq:lstm_update}
\mathbf{c}_t = \mathbf{f}_t * \mathbf{c}_{t-1} + \mathbf{i}_t * \mathbf{g}_t
\end{equation}	
\begin{equation}
\mathbf{h}_t = \mathbf{o}_t * f(\mathbf{c}_t),
\end{equation}	

\noindent where $\mathbf{i}_t,\mathbf{f}_t,\mathbf{o}_t$ are input, output and forget gates at step $t$; $\mathbf{g}_t$ is the vector of cell updates  and $\mathbf{c}_t$ is the updated cell vector used to update the hidden state $\mathbf{h}_t$; \(\sigma\) is the sigmoid function and \(*\) is the element-wise multiplication. 

\newcite{gal2015dropout} proposes to drop the previous hidden state when computing values of gates and updates of the current step, where he samples the dropout mask once for every sequence:

\begin{equation}\label{eq:lstm_candidates_gal}
\begin{pmatrix}
\mathbf{i}_{t} \\
\mathbf{f}_{t} \\
\mathbf{o}_{t} \\
\mathbf{g}_{t} \\ 
\end{pmatrix}
=
\begin{pmatrix}
\sigma(\mathbf{W}_{i} \begin{bmatrix}\mathbf{x}_t , d(\mathbf{h}_{t-1}) \end{bmatrix} + \mathbf{b}_i)\\
\sigma(\mathbf{W}_{f} \begin{bmatrix}\mathbf{x}_t , d(\mathbf{h}_{t-1}) \end{bmatrix} + \mathbf{b}_f) \\
\sigma(\mathbf{W}_{o} \begin{bmatrix}\mathbf{x}_t , d(\mathbf{h}_{t-1}) \end{bmatrix} + \mathbf{b}_o) \\
f(\mathbf{W}_{g} \begin{bmatrix}\mathbf{x}_t , d(\mathbf{h}_{t-1}) \end{bmatrix} + \mathbf{b}_g)
\end{pmatrix}
\end{equation}

\newcite{moon2015rnndrop} propose to apply dropout directly to the cell values and use per-sequence sampling as well:

\begin{equation}\label{eq:lstm_cell_drop}
\mathbf{c}_t = d(\mathbf{f}_t * \mathbf{c}_{t-1} + \mathbf{i}_t * \mathbf{g}_t)
\end{equation}	

In contrast to dropout techniques proposed by \newcite{gal2015dropout} and \newcite{moon2015rnndrop}, we propose to apply dropout to the \textit{cell update} vector $\mathbf{g}_t$ as follows:

\begin{equation}\label{eq:lstm_cand_drop}
\mathbf{c}_t = \mathbf{f}_t * \mathbf{c}_{t-1} + \mathbf{i}_t * d(\mathbf{g}_t)
\end{equation}

Different from methods of \cite{moon2015rnndrop,gal2015dropout}, our approach does not require sampling of the dropout masks once for every training sequence. On the contrary, as we will show in Section \ref{sec:exp}, networks trained with a dropout mask sampled per-step achieve results that are at least as good and often better than per-sequence sampling. Figure \ref{fig:networks} shows differences between approaches to dropout.

The approach of \cite{gal2015dropout} differs from ours in the overall strategy -- they consider network's hidden state as input to subnetworks that compute gate values and cell updates and the purpose of dropout is to regularize these subnetworks. Our approach considers the architecture as a whole with the hidden state as its key part and regularize the whole network. The approach of \cite{moon2015rnndrop} on the other hand is seemingly similar to ours. In Section~\ref{subsec:discussion} we argue that our method is a more principled way to drop recurrent connections in gated architectures.

It should be noted that while being different, the three discussed dropout schemes are not mutually exclusive. It is in general possible to combine our approach and the other two. We expect the merge of our scheme and that of \cite{gal2015dropout} to hold the biggest potential. The relations between recurrent dropout schemes are however out of scope of this paper and we rather focus on studying the relationships of different dropout approaches with the conventional forward dropout.




\noindent{\textbf{Gated Recurrent Unit}} (GRU) networks are a recently introduced variant of a recurrent network with hidden state protected by gates~\cite{DBLP:journals/corr/ChoMBB14}. Different from LSTMs, GRU networks use only two gates $\mathbf{r}_t$ and $\mathbf{z}_t$ to update the cell's hidden state $\mathbf{h}_t$:

\begin{equation}
\begin{pmatrix}
\mathbf{z}_{t} \\
\mathbf{r}_{t} \\
\end{pmatrix}
=
\begin{pmatrix}
\sigma(\mathbf{W}_{z} \begin{bmatrix}\mathbf{x}_t , \mathbf{h}_{t-1} \end{bmatrix} + \mathbf{b}_z)\\
\sigma(\mathbf{W}_{r} \begin{bmatrix}\mathbf{x}_t , \mathbf{h}_{t-1} \end{bmatrix} + \mathbf{b}_r) \\
\end{pmatrix}
\end{equation}
\begin{equation}\label{eq:gru_candidates}
\mathbf{g}_t = f(\mathbf{W}_{g} \begin{bmatrix}\mathbf{x}_t , \mathbf{r}_t * \mathbf{h}_{t-1} \end{bmatrix} + \mathbf{b}_g)
\end{equation}	
\begin{equation}\label{eq:gru_updates}
\mathbf{h}_t = (1 - \mathbf{z}_t ) * \mathbf{h}_{t-1} + \mathbf{z}_t * \mathbf{g}_t
\end{equation}


Similarly to the LSTMs, we propoose to apply dropout to the hidden state updates vector $\mathbf{g}_t$:
\begin{equation}\label{eq:gru_cand_drop}
\mathbf{h}_t = (1 - \mathbf{z}_t ) * \mathbf{h}_{t-1} + \mathbf{z}_t * d(\mathbf{g}_t)
\end{equation}
\label{subsec:discussion}

We found that an intuitive idea to drop previous hidden states directly, as proposed in \newcite{moon2015rnndrop}, produces mixed results. We have observed that it helps the network to generalize better when not coupled with the forward dropout, but is usually no longer beneficial when used together with a regular forward dropout. 


The problem is caused by the scaling of neuron activations during inference. Consider the hidden state update rule in the test phase of an LSTM network. For clarity, we assume every gate to be equal to 1:
\begin{equation}
	\mathbf{h}_t = (\mathbf{h}_{t-1} + \mathbf{g}_t)p ,
\end{equation}
\noindent where \(\mathbf{g}_t\) are update vectors computed by Eq.~\ref{eq:lstm_candidates} and \(p\) is the probability to not drop a neuron.  As \(h_{t-1}\) was, in turn, computed using the same rule, we can rewrite this equation as:
\begin{equation}
\mathbf{h}_t = ((\mathbf{h}_{t-2} + \mathbf{g}_{t-1})p + \mathbf{g}_t)p
\end{equation}
Recursively expanding \(\mathbf{h}\) for every timestep results in the following equation: 
\begin{equation}
\label{eq:big_sum}
\mathbf{h}_t = ((((\mathbf{h}_{0} + \mathbf{g}_{0})p + \mathbf{g}_1)p + ...)p + \mathbf{g}_t)p
\end{equation}
Pushing \(p\) inside parenthesis, Eq.~\ref{eq:big_sum} can be written as:
\begin{equation}\label{eq:drop_test_sum_1}
\mathbf{h}_t = p^{t+1} \mathbf{h}_0 + \sum_{i=0}^{t}p^{t-i+1}\mathbf{g}_i
\end{equation}
Since \(p\) is a value between zero and one, sum components that are far away in the past are multiplied by a very low value and are effectively removed from the summation. Thus, even though the network is able to learn long-term dependencies, it is not capable of exploiting them during test phase. Note that our assumption of all gates being equal to 1 helps the network to preserve hidden state, since in a real network gate values lie within (0, 1) interval. In practice trained networks tend to saturate gate values \cite{karpathyJL15visualizing} what makes gates to behave as binary switches. The fact that \newcite{moon2015rnndrop} have achieved an improvement can be explained by the experimentation domain. \newcite{DBLP:journals/corr/LeJH15} have proposed a simple yet effective way to initialize vanilla RNNs and reported that they have achieved a good result in the Speech Recognition domain while having an effect similar to the one caused by Eq.~\ref{eq:drop_test_sum_1}. One can reduce the influence of this effect by selecting a low dropout rate. This solution however is partial, since it only increases the number of steps required to completely forget past history and does not remove the problem completely. 

One important note is that the dropout function from Eq.~\ref{eq:dropout} can be implemented as:
\begin{equation}
d(\mathbf{x}) = \begin{dcases}
mask * \mathbf{x} / p,& \text{if train phase}\\
\mathbf{x}  		     & \text{otherwise}
\end{dcases}
\end{equation}
In this case the above argument holds as well, but instead of observing exponentially decreasing hidden states during testing, we will observe exponentially increasing values of hidden states during training. 

Our approach addresses the problem discussed previously by dropping the update vectors \(\mathbf{g}\). Since we drop only candidates, we do not scale the hidden state directly. This allows for solving the scaling issue, as Eq.~\ref{eq:drop_test_sum_1} becomes:
\begin{equation}\label{eq:drop_test_sum_2}
\mathbf{h}_t = p\mathbf{h}_0 + \sum_{i=0}^{t}p\ \mathbf{g}_i = p\mathbf{h}_0 + p\sum_{i=0}^{t}\mathbf{g}_i
\end{equation}
Moreover, since we only drop differences that are added to the network's hidden state at each time-step, this dropout scheme allows us to use per-step mask sampling while still being able to learn long-term dependencies. Thus, our approach allows to freely apply dropout in the recurrent connections of a gated network without hindering its ability to process long-term relationships.

We note that the discussed problem does not affect vanilla RNNs because they overwrite their hidden state at every timestep. Lastly, the approach of \newcite{gal2015dropout} is not affected by the issue as well.
\section{Experiments}
\label{sec:exp}
First, we empirically demonstrate the issues linked to memory loss when using various dropout techniques in recurrent nets (see Sec.~\ref{subsec:discussion}). For this purpose we experiment with training LSTM networks on one of the synthetic tasks from~\cite{Hochreiter:1997:LSM:1246443.1246450}, specifically the Temporal Order task. We then validate the effectiveness of our \textit{recurrent dropout} when applied to vanilla RNNs, LSTMs and GRUs on three diverse public benchmarks: Language Modelling, Named Entity Recognition, and Twitter Sentiment classification. 

\subsection{Synthetic Task}
\label{subsec:temp_order}
\noindent{\textbf{Data.}} In this task the input sequences are generated as follows: all but two elements in a sequence are drawn randomly from \{C, D\} and the remaining two symbols from \{A, B\}. Symbols from \{A, B\} can appear at any position in the sequence. The task is to classify a sequence into one of four classes (\{AA, AB, BA, BB\}) based on the order of the symbols. 
We generate data so that every sequence is split into three parts with the same size and emit one meaningful symbol in first and second parts of a sequence. The prediction is taken after the full sequence has been processed. We use two modes in our experiments: \texttt{Short} with sequences of length 15 and \texttt{Medium} with sequences of length 30.

\noindent{\textbf{Setup.}} We use LSTM with one layer that contains 256 hidden units and \textit{recurrent dropout} with 0.5 strength.  Network is trained by SGD with a learning rate of 0.1 for 5k epochs. The networks are trained on 200 mini-batches with 32 sequences and tested on 10k sequences.

\begin{table*}
	\begin{center}
		\begin{tabular}{c|rr|rr|rr|rr}
			\toprule
			       \multirow{3}{*}{Sampling}         & \multicolumn{4}{c|}{\newcite{moon2015rnndrop}}  & \multicolumn{4}{c}{\newcite{gal2015dropout}; Ours}        \\
			                                         & \multicolumn{2}{l|}{short sequences}           & \multicolumn{2}{l|}{medium sequences}          & \multicolumn{2}{l|}{short sequences}         & \multicolumn{2}{l}{medium sequences} \\
			                                         & Train &                                   Test & Train &                                   Test & Train &                                  Test & Train &                         Test \\ \midrule
			                per-step                 & 100\% &                                  100\% &  25\% &                                   25\% & 100\% &                                 100\% & 100\% &                        100\% \\
			              per-sequence               & 100\% &                                   25\% & 100\% &                         \textless 25\% & 100\% &                                 100\% & 100\% &                        100\% \\ \bottomrule
		\end{tabular}
	\end{center}
	\caption{Accuracies on the Temporal Order task.}
	\label{table:order_results}
\end{table*}

\begin{table*}[t]
	\begin{center}
		\begin{tabular}{lc|ll|ll|ll}
			\toprule
			\multirow{2}{*}{Dropout rate} & \multirow{2}{*}{Sampling} & \multicolumn{2}{c|}{\newcite{moon2015rnndrop}} & \multicolumn{2}{c|}{\newcite{gal2015dropout}} & \multicolumn{2}{c}{Ours} \\
			&                           & Valid &             Test & Valid &            Test & Valid &             Test \\ \midrule
			0.0                           &            --             & 130.0 &            125.2 & 130.0 &           125.2 & 130.0 &            125.2 \\
			0.25                          &         per-step          & \textbf{113.0} &            \textbf{108.7} & 119.8 &           114.2 & 106.1 &            100.0 \\
			0.5                           &         per-step          & 124.0 &            116.5 & \textbf{118.3} &           \textbf{112.5} & 102.8 &             98.0 \\
			0.25                          &       per-sequence        & 121.0 &            113.0 & 120.5 &           114.0 & 106.3 &            100.7 \\
			0.5                           &       per-sequence        & 137.7 &            126.2 & 125.2 &           117.9 & \textbf{103.2} &             \textbf{96.8} \\ \midrule\midrule
			0.0                           &            --             &  \textbf{94.1} &             \textbf{89.5} &  94.1 &            89.5 &  94.1 &             89.5 \\
			0.25                          &         per-step          & 113.5 &            105.8 &  \textbf{92.9} &            \textbf{88.4} &  \textbf{91.6} &             \textbf{87.0} \\
			0.5                           &         per-step          & 140.6 &            130.1 &  98.6 &            92.5 & 100.6 &             95.5 \\
			0.25                          &       per-sequence        & 105.7 &             99.9 &  94.5 &            89.7 &  92.4 &             87.6 \\
			0.5                           &       per-sequence        & 125.4 &            117.4 &  98.4 &            92.5 & 107.8 &            101.8 \\ \bottomrule
		\end{tabular}
	\end{center}
	\caption{Perplexity scores of the LSTM network on word level Language Modeling task (lower is better). Upper and lower parts of the table report results without and with forward dropout respectively. Networks with forward dropout use 0.2 and 0.5 dropout rates in input and output connections respectively. Values in bold show best results for each of the recurrent dropout schemes with and without forward dropout.}
	\label{table:lm_results}
\end{table*}

\noindent{\textbf{Results.}} Table \ref{table:order_results} reports the results on the Temporal Order task when \textit{recurrent dropout} is applied using our method and methods from \cite{moon2015rnndrop} and ~\cite{gal2015dropout}. Using dropout from \cite{moon2015rnndrop} with per-sequence sampling, networks are able to discover the long-term dependency, but fail to use it on the test set due to the scaling issue. Interestingly, in \texttt{Medium} case results on the test set are worse than random. Networks trained with per-step sampling exhibit different behaviour: in \texttt{Short} case they are capable of capturing the temporal dependency and generalizing to the test set, but require 10-20 times more iterations to do so. In \texttt{Medium} case these networks do not fit into the allocated number of iterations. This suggests that applying dropout to hidden states as suggested in \cite{moon2015rnndrop} corrupts memory cells hindering the long-term memory capacity of LSTMs.

In contrast, using our \textit{recurrent dropout} methods, networks are able to solve the problem in all cases. We have also ran the same experiments for longer sequences, but found that the results are equivalent to the \texttt{Medium} case. We also note that the approach of \cite{gal2015dropout} does not seem to exhibit the memory loss problem.

\subsection{Word Level Language Modeling}
\label{subsec:lm_exp}
\noindent{\textbf{Data.}} Following~\newcite{MikolovPTB} we use the Penn Treebank Corpus to train our Language Modeling (LM) models. The dataset contains approximately 1 million words and comes with pre-defined training, validation and test splits, and a vocabulary of 10k words.

\noindent{\textbf{Setup.}}
In our LM experiments we use recurrent networks with a single layer with 256 cells. Network parameters were initialized uniformly in [-0.05, 0.05]. For training, we use plain SGD with batch size 32 with the maximum norm gradient clipping \cite{DBLP:conf/icml/PascanuMB13}. Learning rate, clipping threshold and number of Backpropagation Through Time (BPTT) steps were set to 1, 10 and 35 respectively. For the learning rate decay we use the following strategy: if the validation error does not decrease after each epoch, we divide the learning rate by 1.5. The aforementioned choices were largely guided by the work of~\newcite{DBLP:journals/corr/MikolovJCMR14}. To ease reproducibility of our results on the LM and synthetic tasks, we have released the source code of our experiments\footnote{https://github.com/stas-semeniuta/drop-rnn}.

\noindent{\textbf{Results.}} Table \ref{table:lm_results} reports the results for LSTM networks. We also present results when the dropout is applied directly to hidden states as in \cite{moon2015rnndrop} and results of networks trained with the dropout scheme of \cite{gal2015dropout}. We make the following observations: (i) our approach shows better results than the alternatives; (ii) per-step mask sampling is better when dropping hidden state directly; (iii) on this task our method using per-step sampling seems to yield results similar to per-sequence sampling; (iv) in this case forward dropout yields better results than any of the three recurrent dropouts; and finally (v) both our approach and that of \cite{gal2015dropout} are effective when combined with the forward dropout, though ours is more effective.

We make the following observations: (i) dropping hidden state updates yields better results than dropping hidden states; (ii) per-step mask sampling is better when dropping hidden state directly; (iii) contrary to our expectations, when we apply dropout to hidden state updates per-step sampling seems to yield results similar to per-sequence sampling; (iv) applying dropout to hidden state \textit{updates} rather than hidden states in some cases leads to a perplexity decrease by more than 30 points; and finally (v) our approach is effective even when combined with the forward dropout -- for LSTMs we are able to bring down perplexity on the validation set from 130 to 91.6.

\begin{figure}
	\centering
	\centering\includegraphics[width=0.5\textwidth]{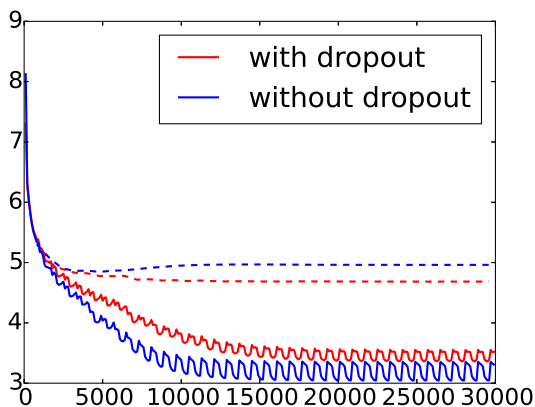}
	\caption{Learning curves of LSTM networks when training without and with 0.25 per-step recurrent dropout. Solid and dashed lines show training and validation errors respectively. Best viewed in color.}
	\label{fig:learning_curves}
\end{figure}

To demonstrate the effect of our approach on the learning process, we also present learning curves of LSTM networks trained with and without \textit{recurrent dropout} (Fig.~\ref{fig:learning_curves}). Models trained using our \textit{recurrent dropout} scheme have slower convergence than models without dropout and usually have larger training error and lower validation errors. This behaviour is consistent with what is expected from a regularizer and is similar to the effect of the feed-forward dropout applied to non-recurrent networks~\cite{hinton_dropout}.

\subsection{Character Level Language Modeling}

\begin{table*}[t]
	\begin{center}
		\begin{tabular}{lc|rr|ll|ll}
			\toprule
			\multirow{2}{*}{Dropout rate} & \multirow{2}{*}{Sampling} & \multicolumn{2}{c|}{\newcite{moon2015rnndrop}} & \multicolumn{2}{c|}{\newcite{gal2015dropout}} & \multicolumn{2}{c}{Ours} \\
			&                           & Valid &             Test & Valid &            Test & Valid &             Test \\ \midrule
			0.0                           &            --             &  1.460     &     1.457             &  1.460     &   1.457    &  1.460     & 1.457 \\
			0.25                          &         per-step          &  1.435     &     1.394             &  1.345     &   1.308    &  \textbf{1.338}     & \textbf{1.301} \\
			0.5                           &         per-step          &  1.610     &     1.561             &  1.387     &   1.348    &  1.355     & 1.316 \\
			0.25                          &       per-sequence        &  \textbf{1.433}     &     \textbf{1.390}             &  \textbf{1.341}     &   \textbf{1.304}    &  1.356     & 1.319 \\
			0.5                           &       per-sequence        &  1.691     &     1.647             &  1.408     &   1.369    &  1.496     & 1.450 \\ \midrule\midrule
			0.0                           &            --             &  \textbf{1.362}     &     \textbf{1.326}             &  \textbf{1.362}     &   \textbf{1.326}    &  1.362     & 1.326 \\
			0.25                          &         per-step          &  1.471     &     1.428             &  1.381     &   1.344    &  \textbf{1.358}     & \textbf{1.321} \\
			0.5                           &         per-step          &  1.668     &     1.622             &  1.463     &   1.425    &  1.422     & 1.380 \\
			0.25                          &       per-sequence        &  1.455     &     1.413             &  1.387     &   1.348    &  1.403     & 1.363 \\
			0.5                           &       per-sequence        &  1.681     &     1.637             &  1.477     &   1.435    &  1.567     & 1.522 \\ \bottomrule
		\end{tabular}
	\end{center}
	\caption{Bit-per-character scores of the LSTM network on character level Language Modelling task (lower is better). Upper and lower parts of the table report results without and with forward dropout respectively. Networks with forward dropout use 0.2 and 0.5 dropout rates in input and output connections respectively. Values in bold show best results for each of the recurrent dropout schemes with and without forward dropout.}
	\label{table:char_lm_results}
\end{table*}
\noindent{\textbf{Data.}} We train our networks on the dataset described in the previous section. It contains approximately 6 million characters, and a vocabulary of 50 characters. We use the provided partitions train, validation and test partitions. 

\noindent{\textbf{Setup.}} We use networks with 1024 units to solve the character level LM task. The characters are embedded into 256 dimensional space before being processed by the LSTM. All parameters of the networks are initialized uniformly in [-0.01, 0.01]. We train our networks on non-overlapping sequences of 100 characters. The networks are trained with the Adam \cite{DBLP:journals/corr/KingmaB14} algorithm with initial learning rate of 0.001 for 50 epochs. We decrease the learning rate by 0.97 after every epoch starting from epoch 10. To avoid exploding gradints, we use MaxNorm gradient clipping with threshold set to 10.

\noindent{\textbf{Results.}} Results of our experiments are given in Table \ref{table:char_lm_results}. Note that on this task regularizing only the recurrent connections is more beneficial than only the forward ones. In particular, LSTM networks trained with our approach and the approach of \cite{gal2015dropout} yield a lower bit-per-character (bpc) score than those trained with forward dropout onlyWe attribute it to pronounced long term dependencies. In addition, our approach is the only one that improves over baseline LSTM with forward dropout. The overall best result is achieved by a network trained with our dropout with 0.25 dropout rate and per-step sampling, closely followed by network with \newcite{gal2015dropout} dropout.

\subsection{Named Entity Recognition}

\noindent{\textbf{Data.}}
To assess our recurrent Named Entity Recognition (NER) taggers when using \textit{recurrent dropout} we use a public benchmark from CONLL 2003~\cite{ner-shared-task}. The dataset contains approximately 300k words split into train, validation and test partitions. Each word is labeled with either a named entity class it belongs to, such as \texttt{Location} or \texttt{Person}, or as being not named. The majority of words are labeled as not named entities. The vocabulary size is about 22k words.

\noindent{\textbf{Setup.}}
Previous state-of-the-art NER systems have shown the importance of using word context features around entities. Hence, we slightly modify the architecture of our recurrent networks to consume the context around the target word by simply concatenating their embeddings. The size of the context window is fixed to 5 words (the word to be labeled, two words before and two words after). The recurrent layer size is 1024 units. The network inputs include only word embeddings (initialized with pretrained word2vec embeddings~\cite{word2vec_embeddings} and kept static) and capitalization features. For training we use the RMSProp algorithm~\cite{DBLP:journals/corr/DauphinVCB15} with \(\rho\) fixed at 0.9 and a learning rate of 0.01 and multiply the learning rate by 0.99 after every epoch. We also combine our \textit{recurrent dropout} (with \textit{per-sequence mask} sampling) with the conventional forward dropout with the rate 0.2 in input and 0.5 in output connections. Lastly, we found that using \(relu(x)=max(x, 0)\) nonlinearity resulted in higher performance than \(tanh(x)\).

To speed up the training we use a length expansion approach described in~\cite{DBLP:journals/corr/NgHVVMT15}, where training is performed in two stages: (i) we first sample short 5-words input sequences with their contexts and train for 25 epochs; (ii) we fine tune the  network on input 15-words sequences for 10 epochs. We found that further fine tuning on longer sequences yielded negligible improvements. Such strategy allows us to significantly speed up the training when compared to training from scratch on full-length input sentences. We use full sentences for testing.

\begin{table}
	\begin{center}
		\begin{tabular}{cccc}
			\toprule
			Dropout rate &  RNN  & LSTM  &     GRU     \\ 
			\midrule
			\multicolumn{4}{c}{5 word long sequences}  \\ \hline
			    0.0      & 85.60  & 85.32 &     86.00      \\
			    0.25     & 86.48 & 86.42 &    86.70     \\ 
			\midrule
			\multicolumn{4}{c}{15 word long sequences} \\ \hline
			    0.0      & 86.12 & 86.12 &    86.44    \\
			    0.25     & 86.95 & 86.88 &    87.10    \\ 
			\bottomrule
		\end{tabular}
	\end{center}
	\caption{F1 scores (higher is better) on NER task.}
	\label{table:ner_results}
\end{table}

\noindent{\textbf{Results.}}
F1 scores of our taggers are reported in Table~\ref{table:ner_results} when trained on short 5-word and longer 15-word input sequences. We note that the gap between networks trained with and without our dropout scheme is larger for networks trained on shorter sequences. It suggests that dropout in recurrent connections might have an impact on how well a network generalizes to sequences that are longer than the ones used during training. The gain from using \textit{recurrent dropout} is larger for the LSTM network. We have experimented with higher \textit{recurrent dropout} rates, but found that it led to excessive regularization.

\subsection{Twitter Sentiment Analysis}
\begin{table*}
	\begin{center}
		\begin{tabular}{c|c|ccccc}
			\toprule
			Dropout rate & Twitter13 & LiveJournal14 & Twitter15 & Twitter14 & SMS13 & Sarcasm14 \\ 
			\midrule
			                              \multicolumn{7}{c}{RNN}                                \\ \hline
			     0       &   67.54   &     71.20      &   59.35   &   68.90    & 64.51 &   53.58   \\
			    0.25     &   66.35   &     67.70      &   59.91   &   67.76   & 64.46 &   48.74   \\ \hline
			                              \multicolumn{7}{c}{LSTM}                               \\ \hline
			     0       &   67.97   &     69.82     &   57.84   &   67.95   & 61.47 &   53.49   \\
			    0.25     &   69.11   &     71.39     &   61.35   &   68.08   & 65.45 &   53.80    \\ \hline
			                              \multicolumn{7}{c}{GRU}                                \\ \hline
			     0       &   67.09   &     70.80      &   59.07   &   67.02   & 67.11 &   51.01   \\
			    0.25     &   69.04   &     72.10      &   60.34   &   69.65   & 65.73 &   54.77   \\ 
			    \bottomrule
		\end{tabular}
	\end{center}
	\caption{F1 scores (higher is better) on Sentiment Evaluation task}
	\label{table:semeval_results}
\end{table*}

\noindent{\textbf{Data.}} We use Twitter sentiment corpus from SemEval-2015 Task 10 (subtask B)~\cite{rosenthal-EtAl:2015:SemEval}. It contains 15k labeled tweets split into training and validation partitions. The total number of words is approximately 330k and the vocabulary size is 22k. The task consists of classifying a tweet into three classes: \texttt{positive}, \texttt{neutral}, and \texttt{negative}. Performance of a classifier is measured by the average of F1 scores of \texttt{positive} and \texttt{negative} classes. We evaluate our models on a number of datasets that were used for benchmarking during the last years. 

\noindent{\textbf{Setup.}} We use recurrent networks in the standard sequence labeling manner - we input words to a network one by one and take the label at the last step. Similarly to \cite{DBLP:conf/sigir/SeverynM15a}, we use 1 million of weakly labeled tweets to pre-train our networks. We use networks composed of 500 neurons in all cases. Our models are trained with the RMSProp algorithm with  a learning rate of 0.001. We use our \textit{recurrent dropout}  regularization with \textit{per-step} mask sampling. All the other settings are equivalent to the ones used in the NER task.

\noindent{\textbf{Results.}} The results of these experiments are presented in Table \ref{table:semeval_results}. Note that in this case our algorithm decreases the performance of the vanilla RNNs while this is not the case for LSTM and GRU networks. This is due to the nature of the problem: differently from LM and NER tasks, a network needs to aggregate information over a long sequence. Vanilla RNNs notoriously have difficulties with this and our dropout scheme impairs their ability to remember even further. The best result over most of the datasets is achieved by the GRU network with \textit{recurrent dropout}. The only exception is the Twitter2015 dataset, where the LSTM network shows better results.

\section{Conclusions}
\label{sec:conclusions}

This paper presents a novel \textit{recurrent dropout} method specifically tailored to the gated recurrent neural networks. Our approach is easy to implement and is even more effective when combined with conventional forward dropout. We have shown that for LSTMs and GRUs applying dropout to arbitrary cell vectors results in suboptimal performance. We discuss in detail the cause of this effect and propose a simple solution to overcome it. The effectiveness of our approach is verified on three different public NLP benchmarks. 

Our findings along with our empirical results allow us to answer the questions posed in Section~\ref{sec:intro}: 
i) while is straight-forward to use dropout in vanilla RNNs due to their strong similarity with the feed-forward architectures, its application to LSTM networks is not so straightforward. We demonstrate that \textit{recurrent dropout} is most effective when applied to \textit{hidden state update} vectors in LSTMs rather than to \textit{hidden states};
(ii) we observe an improvement in the network's performance when our \textit{recurrent dropout} is coupled with the standard forward dropout, though the extent of this improvement depends on the values of dropout rates;
(iii) contrary to our expectations, networks trained with per-step and per-sequence mask sampling produce similar results when using our \textit{recurrent dropout} method, both being better than the dropout scheme proposed by \newcite{moon2015rnndrop}.

While our experimental results show that applying \textit{recurrent dropout} method leads to significant improvements across various NLP benchmarks (especially when combined with conventional forward dropout), its benefits for other tasks, e.g., sequence-to-sequence prediction, or other domains, e.g., Speech Recognition, remain unexplored. We leave it as our future work. 


\section*{Acknowledgments}

This project has received funding from the European Union's Framework Programme for Research and Innovation HORIZON 2020 (2014-2020) under the Marie Skłodowska-Curie Agreement No. 641805. Stanislau Semeniuta thanks the support from Pattern Recognition Company GmbH. We gratefully acknowledge the support of NVIDIA Corporation with the donation of the Titan X GPU used for this research.

\bibliographystyle{acl2016}
\bibliography{acl2015}

\end{document}